\documentclass[
]{ceurart}

\usepackage{listings}
\usepackage{graphicx}
\usepackage{subcaption}
\usepackage{natbib}

\begin{document}

\copyrightyear{2023}
\copyrightclause{Copyright for this paper by its authors.
  Use permitted under Creative Commons License Attribution 4.0
  International (CC BY 4.0).}

\conference{IberLEF 2023, September 2023, Jaén, Spain}

\title{Automated Text Identification Using CNN and Training Dynamics}

\author[1]{Claudiu, Creanga}
\author[1]{Liviu Petrisor, Dinu}
\address[1]{University of Bucharest, Faculty of Mathematics and Computer Science, Bucharest, Romania}

\begin{abstract}
    We used Data Maps to model and characterize the AuTexTification dataset. This provides insights about the behaviour of individual samples during training across epochs (training dynamics). We characterized the samples across 3 dimensions: confidence, variability and correctness. This shows the presence of 3 regions: easy-to-learn, ambiguous and hard-to-learn examples. We used a classic CNN architecture and found out that training the model only on a subset of ambiguous examples improves the model's out-of-distribution generalization.  
\end{abstract}

\begin{keywords}
  data maps \sep
  AuTexTification \sep
  CNN \sep
  CEUR-WS
\end{keywords}

\maketitle

\section{Introduction}

The AuTexTification challenge \cite{autextification} aims to tackle the complex task of distinguishing between human and AI generated text. As AI continues to evolve and generate increasingly sophisticated content, it becomes crucial to develop robust methods and algorithms that can accurately discern between human and AI-written texts. By developing tools that are able to differentiate between human and AI-generated content, we can help identify instances of misinformation, deepfakes or propaganda and also help create a space where AI can be used ethically.

One of the issues that we have in AI, and in NLP in particular, is that often we create models which obtain a high score for our training and test datasets, but then in practice the models don't perform that well \cite{Linzen20}. In other words, out-of-distribution (ODD) performance, doesn't match in-distribution (ID) performance. In order to encourage models that generalize better, the AuTexTification dataset used data from three domains for training, and from two different domains for testing. 

To improve the generalization of our model we use the approach presented by Swayamdipta \cite{swayamdipta-etal-2020-dataset} which draws analogies from cartography to look at the dynamics of individual samples. The tool they provided is working only on 5 NLI (Natural Language Inference) datasets, but with some modifications we could use it with our dataset. Data maps is a tool that looks at the predictions for individual examples across the epochs. With it we can assess for which examples the model does great and which examples are harder to learn. Training the model on different categories of examples and then testing it on the test dataset (which is from two different writing styles) will enable us to tell which kind of examples help the model generalise better on ODD examples. 

\begin{figure}
  \centering
  \includegraphics[width=\linewidth]{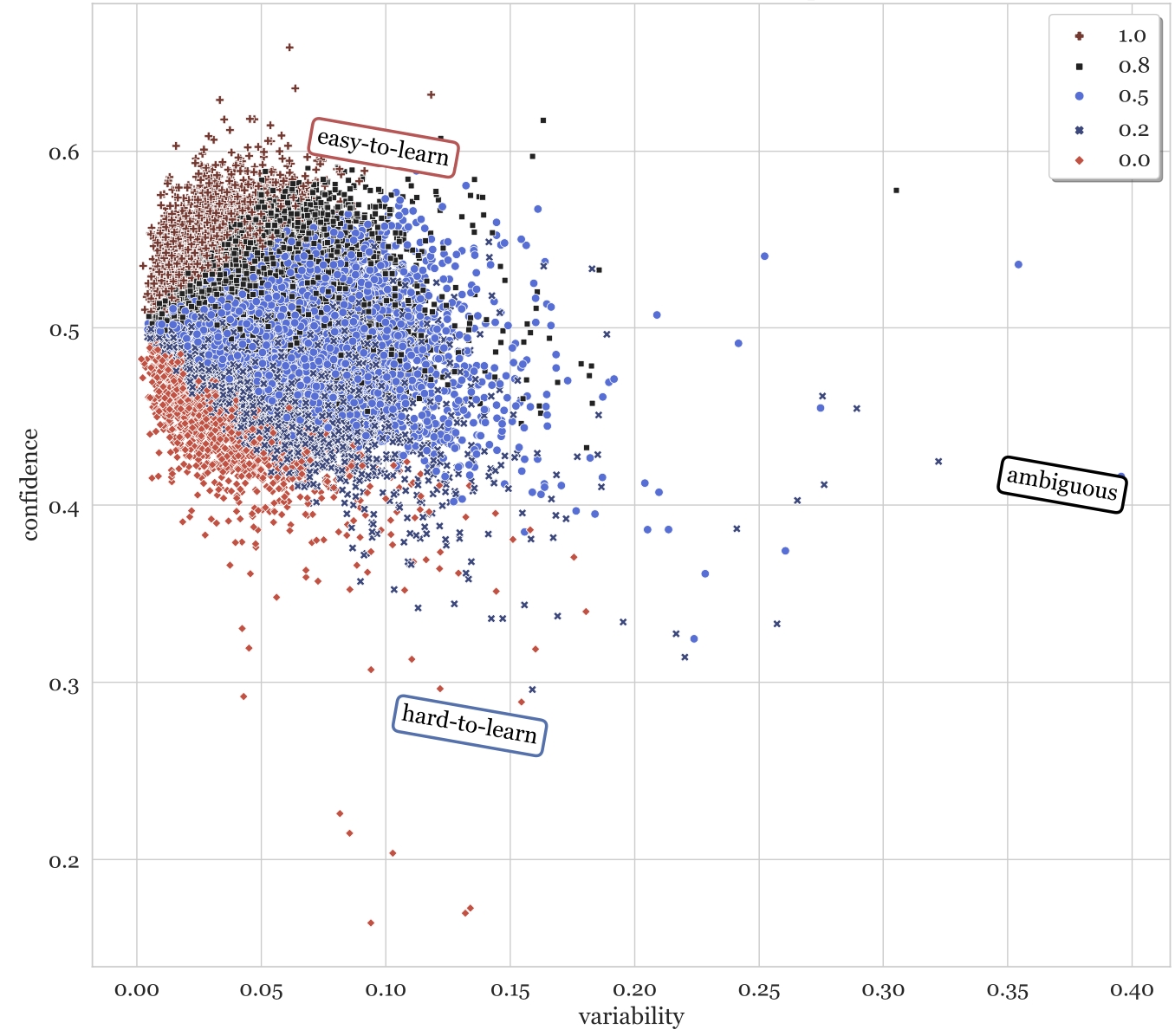}
  \caption{Data map for AuTexTification train dataset, based on our CNN classifier presented in \hyperref[sec:Model]{section 3}
.}
\end{figure}

\section{Training Dynamics}

In Figure 1 we present the data map for the AuTexTification train dataset that was built using our CNN classifier. The y-axis shows the confidence of the model in its prediction and the x-axis shows the variability of the predictions. The different shapes of the glyphs mark the correctness of the prediction. We run the model for 5 epochs and we expressed this through 5 different shapes. One of them, the plus brown sign (+) that we can see is present abundantly in the top left side of the plot, denotes the predictions for the same example that was correct across all 5 epochs (1.0 performance). In the middle we see mostly blue circles, which are predictions that were right about 50\% of the time (2 or 3 epochs). In the bottom we see mostly predictions for examples that were never right, denoted with a red diamond (0.0). We can distinguish 3 regions: easy-to-learn examples in the top left where we have low variability and high confidence, ambiguous examples in the middle to the right where we have high variability and hard-to-learn examples on the bottom left where we have low confidence and low variability.

\begin{table}
  \caption{Mapping of the three regions along the 3 axes: confidence, variability and correctness.}
  \label{tab:regions}
  \begin{tabular}{cccc}
    \toprule
    Region & Confidence & Variability & Correctness\\
    \midrule
    \texttt easy-to-learn & high & low & high \\
    \texttt ambiguous & average & average & average\\
    \texttt hard-to-learn & low & high & low \\
    \bottomrule
  \end{tabular}
\end{table}

The 3 regions with their specifics are presented in Table 1 above. We gave each example in the training dataset an internal ID that we could follow it through our training. We trained our model for 5 epochs and in each epoch we saved the logits individually for each examples. The logits showed the unnormalized scores for the 2 classes (human and generated) that the model learned. If these scores would diverge greatly across the epochs and the model will classify the example as being human in some epochs and as being generated in other epochs, then the example has high variability and will be in the ambiguous region. Otherwise, if the scores would not diverge a lot across the epochs and the model will classify the example correctly in most epochs, the example will have low variability and high confidence and will be in the easy-to-learn region. 

Below in Figure 2 we can see the density plots for our measures. Most of our examples have a confidence between 0.3 and 0.7, which suggests that the model is not very confident in most of it's predictions. This suggests that either the dataset is too challenging for our model's architecture or we need better feature engineering to help the model learn. The correctness plot also shows that the model needs to be improved, ideally we would have the biggest bar at 1.0, but instead we have the biggest bar at 0.5.

\begin{figure}[htbp]
  \centering
    \begin{subfigure}[b]{0.3\textwidth}
    \includegraphics[width=\textwidth]{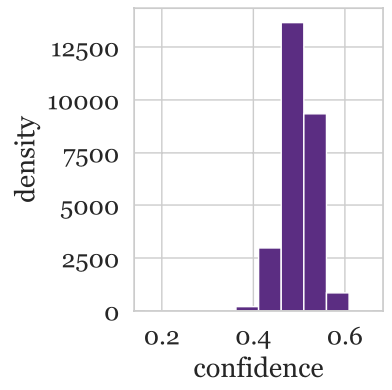}
    \caption{Density plot for confidence.}
    \label{fig:figure1}
  \end{subfigure}
  \hfill
  \begin{subfigure}[b]{0.3\textwidth}
    \includegraphics[width=\textwidth]{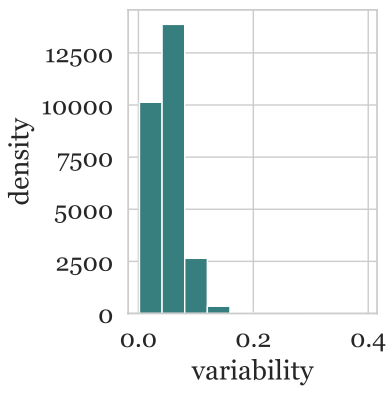}
    \caption{Density plot for variability.}
    \label{fig:figure2}
  \end{subfigure}
  \hfill
  \begin{subfigure}[b]{0.3\textwidth}
    \includegraphics[width=\textwidth]{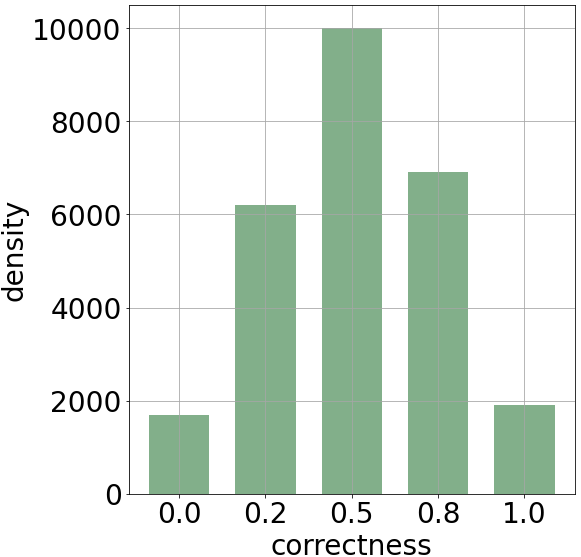}
    \caption{Density plot for correctness.}
    \label{fig:figure3}
  \end{subfigure}
  \caption{Density plots for the three dimensions.}
  \label{fig:three_figures}
\end{figure}

\section{Model}
\label{sec:Model} 
Several papers \cite{mitrović2023} suggest that there are a couple of features that we can look at in order to help our models detect AI generated text. The generated text is more impersonal, uses less pronouns, is overly polite and does not express feelings. 

\begin{figure}
  \centering
  \includegraphics[width=\linewidth]{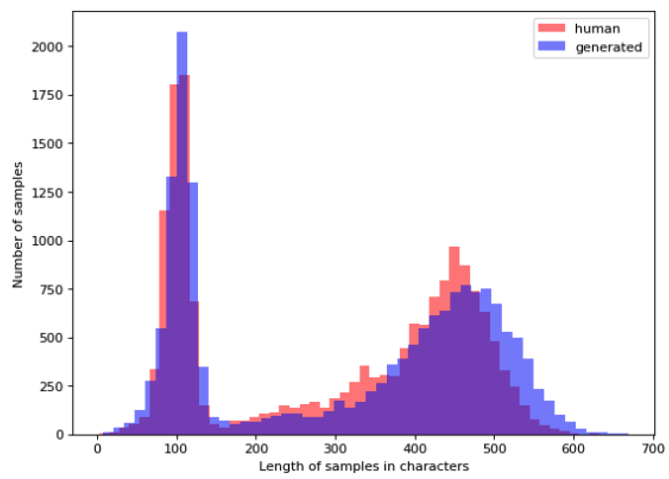}
  \caption{Length distribution for human and generated text. On the x-axis we have the length of characters in each example and on the y-axis the number of examples. The two classes don't differ in the length of characters.}
\end{figure}

\begin{table}
  \caption{Possible features used to distinguish between human and generated sentences: count of pronouns, mean positive and mean negative Vader scores.}
  \label{tab:regions}
  \begin{tabular}{ccc}
    \toprule
    Feature & Generated & Human\\
    \midrule
    \texttt Count of pronouns & 77717 & 71797 \\
    \texttt Mean positive score & 0.11 & 0.09 \\
    \texttt Mean negative score & 0.03 & 0.04 \\
    \bottomrule
  \end{tabular}
\end{table}

Looking for features that could help the model, as we can see in Table 2 above, in our dataset the count of pronouns is relatively the same. Research has shown that, in general, AI generated text have less pronouns than human text. This proves to not be true in this dataset. The size of the text is also equal (Figure 3). We observe two peaks at around 100 characters and 450 characters which suggests that the different writing styles have different length. Using the Vader sentiment analyser, we observed that the generated text is more positive than the human text. Other features that we looked at and didn't help were TF-IDF (which was expected since TF-IDF cannot capture the nuanced differences between human and machine-generated text).

We chose the following features for our classifier: 
\begin{itemize}
\item Vader sentiment scores;
\item count of misspellings;
\item count of syllables;
\item count of stop words;
\item word frequency;
\item Google news Word2Vec which gave us 300-dimensional vectors.
\end{itemize}

These features were fed into a CNN model with 5 Conv1D layers followed by a BatchNormalization layer and a Dropout layer of 0.3. At the end of our model we had 3 fully connected layers. 

\section{Results}

By training our CNN model on the whole training dataset and then testing it on the provided test dataset, we obtain an F1 score of 62, which is lower than the baseline Logistic Regression. If we train our model only on a certain region we find that the model doesn't converge when we train only on hard-to-learn examples. If we train it only on easy-to-learn examples, the model obtains an F1 score of 58, which is lower than training it on the whole dataset. If we train it only on ambiguous examples we obtained a better score than training it on the whole dataset: 64. We then tried to find the right size of the dataset in order to obtain the best F1 score. We trained the model multiple times on 15\%, 25\%, 45\%, 75\% of the ambiguous examples and found that the model obtains the greatest score (F1 = 66.1) on the test dataset using only 45\% of the ambiguous examples (Table 3).

\begin{table}
  \caption{Our results trying various sample sizes and mixing of samples between different regions.}
  \label{tab:regions}
  \begin{tabular}{cc}
    \toprule
    Percentage of samples & F1 score\\
    \midrule
    \texttt 50\% easy-to-learn samples & 58 \\
    \texttt 15\% ambiguous samples and 15\% easy-to-learn samples & 61 \\
    \texttt 15\% ambiguous samples & 62 \\
    \texttt 25\% ambiguous samples & 64 \\
    \texttt 45\% ambiguous samples & 66 \\
    \texttt 75\% ambiguous samples & 64 \\
    \texttt 45\% ambiguous samples and 25\% easy-to-learn samples & 64 \\
    \texttt 50\% hard-to-learn samples & 0 \\
    \bottomrule
  \end{tabular}
\end{table}

\section{Conclusion}
We showed that we can obtain better accuracy on out of distribution data (the test set composed from 2 separated writing styles) by using just a subset of the ambiguous examples from the training dataset. In the end, our CNN obtained an F1 score of 66 when training on just 28\% of the data compared to 62 when training it on the whole dataset. This shows that the model learns more from some examples and less from others. It also gives a direction when building new datasets to focus more on ambiguous examples, rather than on easy-to-learn or hard-to-learn examples. 


\bibliography{references}

\end{document}